\documentclass[10pt,twocolumn,letterpaper]{article}

\usepackage{iccv}
\usepackage{times}
\usepackage{epsfig}
\usepackage{graphicx}
\usepackage{amsmath}
\usepackage{amssymb}
\usepackage{breqn}
\usepackage{float}
\usepackage{algorithm}
\usepackage{algorithmic}
\usepackage{amsmath}
\usepackage{capt-of}
\usepackage{pdfpages}
\usepackage{enumerate}
\usepackage{footnote}
\usepackage{multirow,xcolor}
\usepackage{makecell}
\usepackage{subcaption}
\usepackage[pagebackref=true,breaklinks=true,letterpaper=true,colorlinks,bookmarks=false]{hyperref}

\iccvfinalcopy 

\def\iccvPaperID{1870} 

\ificcvfinal\fi
\begin{document}

\title{Aggregation via Separation:
Boosting Facial Landmark Detector with Semi-Supervised Style Translation}

\author{Shengju Qian\textsuperscript{1},~~~  Keqiang Sun\textsuperscript{2},~~~ Wayne Wu\textsuperscript{2,3},~~~ Chen Qian\textsuperscript{3},~~~ Jiaya Jia\textsuperscript{1,4} \\
\textsuperscript{1}The Chinese University of Hong Kong~~~ \textsuperscript{2}Tsinghua University \\ \textsuperscript{3}SenseTime Research~~~ \textsuperscript{4}YouTu Lab, Tencent \\
{\tt\small \{sjqian, leojia\}@cse.cuhk.edu.hk, skq17@mails.tsinghua.edu.cn, \{wuwenyan, qianchen\}@sensetime.com}
}

\maketitle
\ificcvfinal\thispagestyle{empty}\fi
\begin{abstract}
  Facial landmark detection, or face alignment, is a fundamental task that has been extensively studied. In this paper, we investigate a new perspective of facial landmark detection and demonstrate it leads to further notable improvement. Given that any face images can be factored into space of style that captures lighting, texture and image environment, and a style-invariant structure space, our key idea is to leverage disentangled style and shape space of each individual to augment existing structures via style translation. With these augmented synthetic samples, our semi-supervised model surprisingly outperforms the fully-supervised one by a large margin. Extensive experiments verify the effectiveness of our idea with state-of-the-art results on WFLW~\cite{LAB}, 300W~\cite{sagonas2013300}, COFW~\cite{burgos2013robust}, and AFLW~\cite{AFLW} datasets. Our proposed structure is general and could be assembled into any face alignment frameworks. The code is made publicly available at \href{https://github.com/thesouthfrog/stylealign}{https://github.com/thesouthfrog/stylealign}.
\end{abstract}

\section{Introduction}

Facial landmark detection is a fundamentally important step in many face applications, such as face recognition~\cite{liu2017sphereface}, 3D face reconstruction~\cite{feng2018joint}, face tracking~\cite{khan2017synergy} and face editing~\cite{thies2016face2face}. Accurate facial landmark localization was intensively studied with impressive progress made in these years. The main streams are learning a robust and discriminative model through effective network structure~\cite{LAB}, usage of geometric information~\cite{bulat2017far,pifa}, and correction of loss functions~\cite{feng2018wing}. 

\begin{figure}[t]
\begin{center}
   \includegraphics[width=\linewidth]{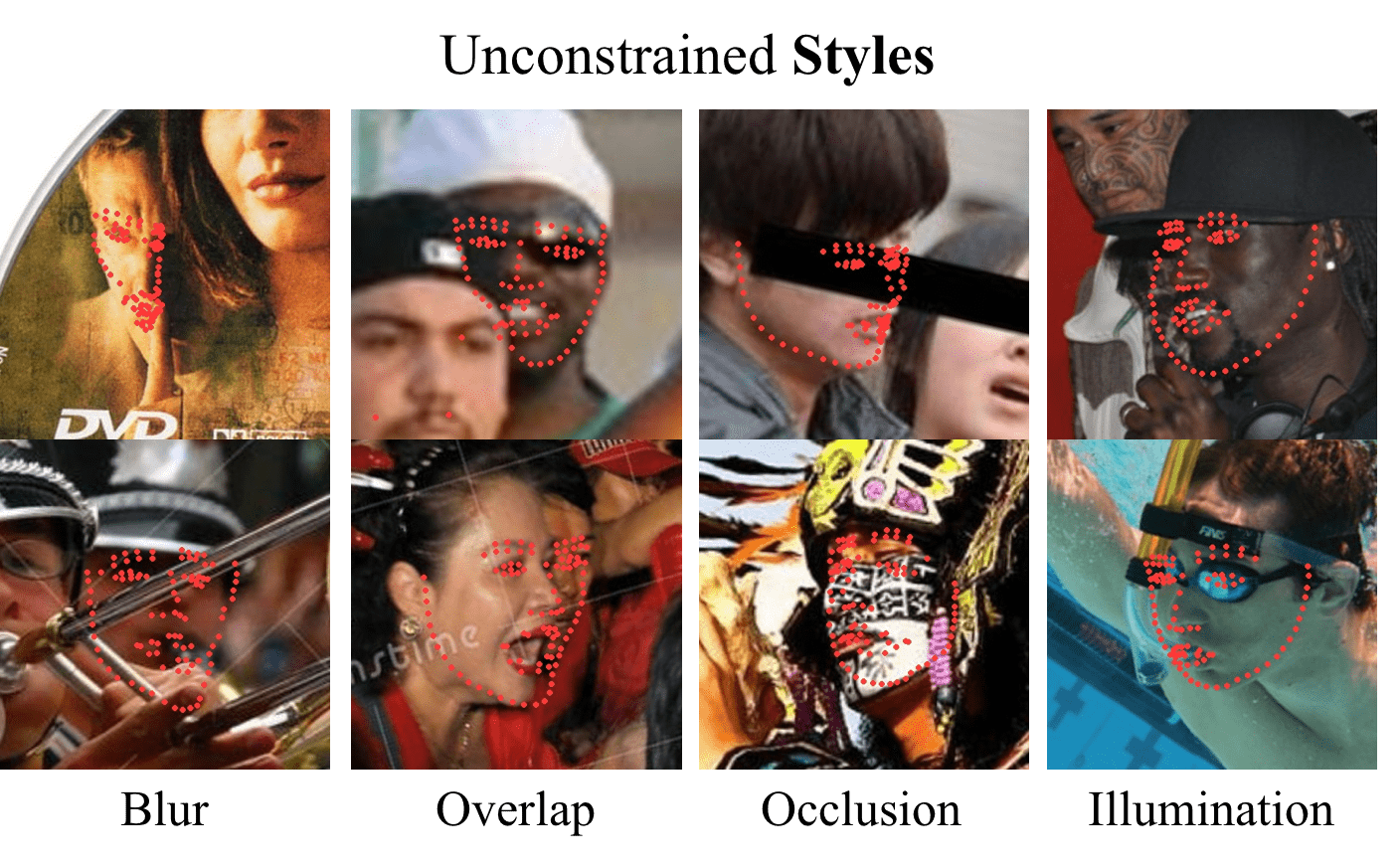}
\end{center}
\vspace{-0.5cm}
   \caption{{Problem in a well-trained facial landmark detector. It is biased towards unconstrained environment factors, including lighting, image quality, and occlusion. We regard these degradations as ``style'' in our analysis.  }}
\label{fig:backcase}
\end{figure}

It is common wisdom now that factors such as variation of expression, pose, shape, and occlusion could greatly affect performance of landmark localization. Almost all prior work aims to alleviate these problems from the perspective of structural characteristics, such as disentangling 3D pose to provide shape constraint~\cite{PCD-CNN}, and utilizing dense boundary information~\cite{LAB}. The influence of ``environment'' still lacks principled discussion beyond structure. Also, considering limited labeled data for this task, how to optimally utilize limited training samples remains unexplored. 

About ``environment" effect, distortion brought by explicit image style variance was observed recently~\cite{SAN}. We instead utilize style transfer~\cite{johnson2016perceptual,gatys2016image} and disentangled representation learning~\cite{tran2017disentangled,chen2016infogan, DRIT, vunet, bvae} to tackle the face alignment problem, since style transfer aims at altering style while preserving content. In practice, image content refers to objects, semantics and sharp edge maps, whereas style could be color and texture. 

Our idea is based on the purpose of facial landmark detection, which is to regress ``facial content'' -- the principal component of facial geometry -- by filtering unconstrained ``styles''. The fundamental difference to define ``style" from that of \cite{SAN} is that we refer it to image background, lighting, quality, existence of glasses, and other factors that prevent detectors from recognizing facial geometry. We note every face image can be decomposed into its facial structure along with a distinctive attribute. It is a natural conjecture that {\it face alignment could be more robust if we augment images only regarding their styles}.

To this end, we propose a new framework to augment training for facial landmark detection without using extra knowledge. Instead of directly generating images, we first map face images into the space of structure and style. To guarantee the disentanglement of these two spaces, we design a conditional variational auto-encoder~\cite{kingma2013auto} model, in which Kullback-Leiber (KL) divergence loss and skip connections are incorporated for compact representation of style and structure respectively. By factoring these features, we perform visual style translation between existing facial geometry. Given existing facial structure, faces with glasses, of poor quality, under blur or strong lighting are {\it re-rendered} from corresponding style, which are used to further train the facial landmark detectors for a rather general and  robust system to recognize facial geometry. 

Our main contribution is as follows.

\begin{enumerate}
    \item We offer a new perspective for facial landmark localization by factoring style and structure. Consequently, a face image is decomposed and rendered from distinctive image style and facial geometry. 
    \item A novel semi-supervised framework based on conditional variational auto-encoder is built upon this new perspective. By disentangling style and structure, our model generates style-augmented images via style translation, further boosting facial landmark detection. 
    \item We propose a new dataset based on AFLW~\cite{AFLW} with new 68-point annotation. It provides challenging benchmark considering large pose variation. 
\end{enumerate}

With extensive experiments on popular benchmark datasets including WFLW~\cite{LAB}, 300W~\cite{sagonas2013300}, COFW~\cite{burgos2013robust} and AFLW~\cite{AFLW}, our approach outperforms previous state-of-the-arts by a large margin. It is general to be incorporated into various frameworks for further performance improvement. Our method also works well under limited training computation resource.

\section{Related Work}

This work has close connection with the areas of facial landmark detection, disentangled representation and self-supervised learning.

\vspace{-0.1in}
\paragraph{Facial Landmark Detection} This area has been extensively studied over past years. Classic parameterized methods, such as active appearance models (AAMs)~\cite{AAM,nonlinearAAM,AAMR,kahraman2007active} and constrained local models (CLMs)~\cite{CLM} provide satisfying results. SDM~\cite{SDM}, cascaded regression, and their variants~\cite{cpm_true, zhu2016unconstrained, CFSS, ESR, burgos2013robust, chen2014joint,SDM,tuzel2016robust,feng2015cascaded} were also proposed. 

Recently, with the power of deep neural networks, regression-based models are able to produce better results. They are mainly divided into two streams of direct coordinate regression~\cite{TCDCN,TSR,MDM,miao2018direct} and heatmap-based regression~\cite{hourglass,CALE,deng2017joint,menpo2017,merget2018robust}. Meanwhile, in~\cite{TCDCN}, auxiliary attributes were used to learn a discriminative representation. Recurrent modules~\cite{MDM,rar,pengrecurrent} were introduced then. Lately, methods improved performance via semi-supervised learning~\cite{semi}. Influence of style variance was also discussed in~\cite{SAN}, where a style aggregated component provides a stationary environment for landmark detector. Our solutions are distinct with definition of ``style", different from prior work. Our solution does not rely on the aggregation architecture, and instead is based on a semi-supervised scheme.

\vspace{-0.1in}
\paragraph{Disentangled Representation} Our work is also related to disentangled representation learning. Disentanglement is necessary to control and further alter the latent information in generated images. Under the unsupervised setting, InfoGAN~\cite{chen2016infogan} and MINE~\cite{MINE} learned disentangled representation by maximizing the mutual information between latent code and data observation. Recently, image-to-image translation~\cite{UNIT,MUNIT,DRIT,cGAN} explored the disentanglement between style and content without supervision. 
In structured tasks such as conditional image synthesis~\cite{poseguided}, keypoints~\cite{vunet,pumarola2018unsupervised} and person mask~\cite{balakrishnan2018synthesizing} were utilized as self-supervision signals to disentangle factors, such as foreground, background and pose information. As our ``style" is more complex while ``content'' is represented by facial geometry, traditional style transfer~\cite{gatys2016image} is inapplicable since it may suffer from structural distortion. In our setting, by leveraging the structure information base on landmarks, our separation component extracts the style factor from each face image. 

\begin{figure*}[htb]
\begin{center}
 \includegraphics[width=1.0\linewidth]{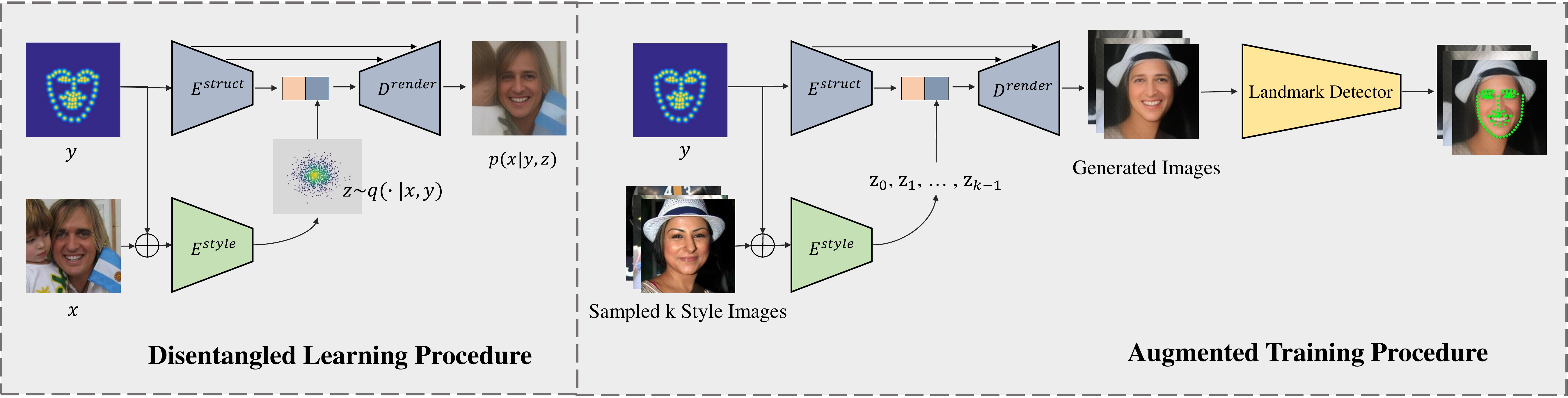}
\end{center}
\vspace{-0.4cm}
   \caption{{Our framework. It consists of two stages. The first stage is to train the network to disentangle face images to style and structure space. At the second stage, style translation is performed to augment training of facial landmark detectors. }}
\label{fig:framework}
\vspace{-0.2cm}
\end{figure*}

\vspace{-0.1in}
\paragraph{Self-Supervised Learning}
Our method also connects to self-supervised learning. The mainstream work, such as~\cite{zamir2018taskonomy}, directly uses image data to provide proxy supervision through multi-task feature learning. Another widely-adopted approach is to use video data~\cite{wang2015unsupervised}. Visual invariance of the same instance could be captured in a consecutive sequence of video frames~\cite{foldiak1991learning,wiskott2002slow,le2011learning,zou2012deep,taylor2010convolutional,stavens2010unsupervised,wang2015unsupervised}. Also, there is work focusing on fixed characteristics of objects from data statistics~\cite{doersch2015unsupervised,zhang2016colorful,zhang2017split,larsson2016learning,larsson2017colorization}, such as image patch level information~\cite{doersch2015unsupervised}. These methods learn visual invariance, which could essentially provide a generalized feature of objects. 

Our landmark localization involves computing the visual invariance. 
But our approach is different from prior self-supervised frameworks. Our goal lies in extracting facial structure and keypoints considering different environment factors, including occlusion, lighting, makeup and so on. Eliminating the influence of style makes it possible to reliably alter or process face structure and accordingly recognize invariant features. It thus better deals with style variation, which commonly exists in natural images.

\section{Proposed Framework}

Our framework consists of two parts. One learns the disentangled representation of facial appearance and structure, while the other can be any facial landmark detectors. As illustrated in Fig.~\ref{fig:framework}, during the first phase, conditional variational auto-encoder is proposed for learning disentangled representation between style and structure. In the second phase, after translating style from other faces, ``stylized'' images with their structures are available for boosting training performance and our style-invariant detectors. 

\subsection{Learning Disentangled Style and Structure }

Given an image $x$, and its corresponding structure $y$. Two essential descriptors of a face image are facial geometry and image style. Facial geometry is represented by labeled landmarks, while style captures all environmental factors that are mostly implicit, as described above. With this setting, if the latent space of style and shape is mostly uncorrelated, using Cartesian product of $z$ and $y$ latent space should capture all variation included in a face image. Therefore, the generator that re-renders a face image based on style and structure can be modeled as $p(x|y, z)$. 

To encode the style and structure information and compute the parametric distribution $p(x|y, z)$, a conditional variational auto-encoder based network, which introduces two encoders, is applied. Our network consists of a structure estimator $E^{struct}$ to encode landmark heatmaps into structure latent space, a style encoder $E^{style}$ that learns the style embedding of images, and a decoder that re-renders the style and structure to image space. 

As landmarks available in this task, the facial geometry is represented by stacking landmarks to heat maps. Our goal therefore becomes inferring disentangled style code $z$ from a face image and its structure by maximizing the conditional likelihood of
\begin{equation}
\begin{split}
	\log p(x|y) = \log \int_z p(x, z|y) dz \ge \mathbb{E}_q[\log \frac{p(x,z|y)}{q(z|x,y)}] \\ = \mathbb{E}_q[\log p(x|z,y)]-D_{KL}[q(z|x,y),p(z|y)].
	\label{eq:likihood}
\end{split}
\end{equation}
In particular, the generator $G^{full}_{\theta}$ contains two encoders and a decoder (renderer), i.e., $E^{style}_{\phi}$, $E^{struct}$ and $D_{render}$, where $G^{full}_{\theta}$ and $E^{style}_{\phi}$ respectively estimate parameters of $p(x|y, z)$ and $q(z|x,y)$. Consequently, the full loss function on learning separating information of style and structure is written as
\begin{equation}
\begin{split}
	\mathcal{L}_{disentangle}(x, \theta, \phi) = - KL (q_{\phi}(z|x, y)) || p_{\theta}(z|y)) \\ + \mathcal{L}_{rec}(x, G^{full}(E^{style}(x,y), E^{struct}(y)).
	\label{eq:loss}
\end{split}
\end{equation}

\begin{figure*}[t!]
	\begin{center}
		\includegraphics[width=\linewidth]{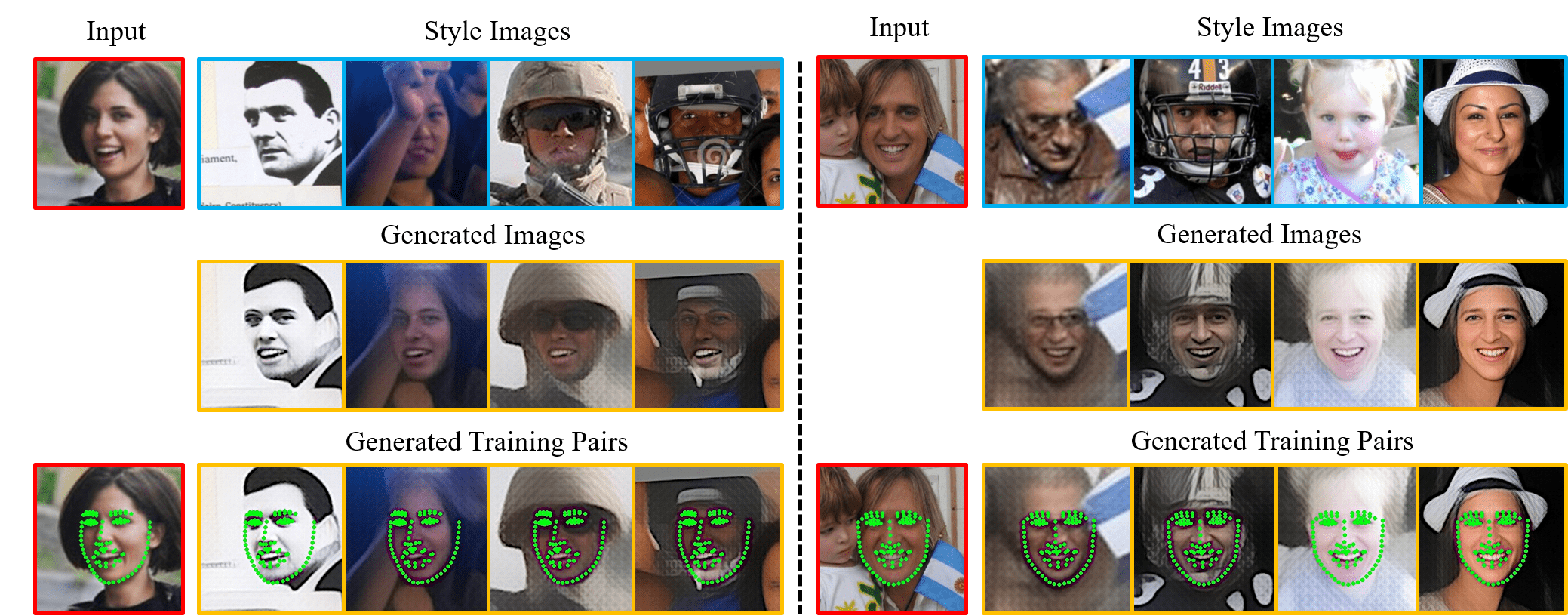}
	\end{center}
	\vspace{-0.3cm}
	\caption{{Visualization of style translation. Given the input images in red， $4$ different styles are provided to perform translation towards input structure. The synthetic images along with input original landmarks are provided to demonstrate the strong coherence of structure. }}
	\vspace{-0.2cm}
	\label{fig:generation}
\end{figure*}

\vspace{-0.12in}
\paragraph{KL-Divergence Loss} Kullback-Leiber (KL) divergence loss severs as a key component in our design to help the encoder to learn decent representation. Basically, the KL-divergence measures the similarity between the variational posterior and prior distribution. In our framework, it is taken as regularization that discourages $E^{style}$ to encode structure-related information. As the prior distribution is commonly assumed to be a unit Gaussian distribution $p\sim N(0,1)$, the learned style feature is regularized to suppress contained structure information through reconstruction. 

The KL-divergence loss limits the distribution range and capacity of the style feature. By fusing inferred style code $z$ with encoded structure representation, sufficient structure information can be obtained from prior through multi-level skip connection. Extra structure encoded in $z$ incurs penalty of the likelihood $p(x|y,z)$ during training with no new information captured. In this way, $E^{style}$ is discouraged from learning structure information that is provided by $E^{struct}$ during training. To better reconstruct the original image, $E^{style}$ is enforced to learn structure-invariant style information. 

\vspace{-0.12in}
\paragraph{Reconstruction Loss} The second term $\mathcal{L}_{rec}$ in Eq.~\eqref{eq:loss} refers to the reconstruction loss in the auto-encoder framework. As widely discussed~\cite{zhu2017unpaired,johnson2016perceptual}, basic pixel-wise $L_1$ or $L_2$ loss cannot model rich information within images well. We instead adopt perceptual loss to capture style information and better visual quality. $\mathcal{L}_{rec}$ is formulated as
\begin{equation}
\begin{split}
	\mathcal{L}_{rec}(x, \theta, \phi) = \sum_{l} ||(\Phi_l(x) - \Phi_l(G^{full}(x,y)) ||_2^2,
	\label{eq:percept_loss}
\end{split}
\end{equation}
where we use VGG-19 network $\Phi$ structure that measures perceptual quality. $l$ indexes the layer of network $\Phi$. 

Since the style definition could be complicated, $E^{style}$ here encodes semantics of the style signal that simulates different types of degradation. It does not have to maintain fine-grained visual details. Besides, to reserve the strong prior on structure information encoded from landmarks $y$, skip connection between $E^{struct}$ and $D^{render}$ is established to avoid landmark inaccuracy through style translation. 

In this design, the model is capable of learning complementary representation of facial geometry and image style. 

\subsection{Augmenting Training via Style Translation}

Disentanglement of structure and style forms a solid foundation for diverse stylized face images under invariant structure prior. 

Given a dataset $X$ that contains $n$ face images with landmarks annotation, each face image $x_i (1\le i \le n)$ within the dataset has its explicit structure denoted by landmark $y_i$, as well as an implicit style code $z_i$ depicted and embedded by $E^{style}$. To perform style translation between two images $x_i$ and $x_j$, we pass their latent style and structure code embedded by $E^{style}$ and $E^{struct}$ to $D^{render}$. To put the style of image $x_j$ on $x_i$'s structure, the stylized synthetic image is denoted as
\begin{equation}
\begin{split}
	x_{ij} = D^{render} (E^{style}(x_j ,y_i),E^{struct}(y_i)).
	\label{eq:render}
\end{split}
\end{equation}
As illustrated in Fig.~\ref{fig:framework}, the first stage of our framework is to train the disentangling components. In the second phase, by augmenting and rendering a given sample $x$ in the original dataset $X$ with styles from random $k$ other faces, we produce $k \times n$ ``stylized'' synthetic face images with respective annotated landmarks. These samples are then fed into training of facial landmark detectors together with the original dataset. Visualization of style translation results is provided in Fig.~\ref{fig:generation}. The input facial geometry is maintained under severe style variation, indicating its potential at augmenting training of facial landmark detectors.

Albeit with cohesive structure, the decoder generally does not re-render perfect-quality images, since the complexity of plentiful style information has been diminished to a parametric Gaussian distribution, confined by its capacity. Also, as discussed before, each face image $x_i$ has its own style. Theoretically, the renderer could synthesize $n^2$ images by rendering each available landmark with any other images' style. To understand how the quantity of stylized synthetic samples helps improve the facial landmark detectors, we analyze the effect of our design in following experiments and ablation study. 

\section{Experiments}

\subsection{Datasets}

\begin{table*}[htb]
\begin{center}
\resizebox{0.95\textwidth}{!}{%
\begin{tabular}{c|c|c|c|c|c|c|c|c} 
\Xhline{1.2pt}
Metric                             & Method                 & Fullset         & Pose~           & ~Expression~    & ~Illumination~  & ~Make-Up~       & ~Occlusion~     & Blur             \\ 
\Xhline{1.1pt}
\multirow{11}{*}{Mean Error (\%)}   & CFSS~\cite{CFSS}  & 9.07            & 21.36           & 10.09           & 8.30            & 8.74            & 11.76           & 9.96             \\
                                   & DVLN~\cite{DVLN}  & 6.08            & 11.54           & 6.78            & 5.73            & 5.98            & 7.33            & 6.88             \\
                                   & LAB~\cite{LAB}                    & 5.27            & 10.24           & 5.51            & 5.23            & 5.15            & 6.79            & 6.32             \\
                                   & SAN~\cite{SAN}                    & 5.22            & 10.39           & 5.71            & 5.19            & 5.49            & 6.83            & 5.80             \\
                                   & WING~\cite{feng2018wing}                   & 5.11            & 8.75            & 5.36            & 4.93            & 5.41            & 6.37            & 5.81             \\ 
                                   & Res-18                 & 6.09            & 10.76           & 6.97            & 5.83            & 6.19            & 7.15            & 6.67             \\
\cline{2-9}
                                   & \textbf{Ours w. Res-18} & 5.25            & 9.10            & 5.83            & 4.93            & 5.47            & 6.26            & 5.86             \\
                                   & \textbf{Ours w. LAB}    & \textbf{4.76}   & \textbf{8.21}   & \textbf{5.14}   & \textbf{4.51}   & \textbf{5.00}   & \textbf{5.76}   & \textbf{5.43}    \\
                                   & \textbf{Ours w. SAN}    & \textbf{4.39}   & \textbf{8.42}   & \textbf{4.68}   & \textbf{4.24}   & \textbf{4.37}   & \textbf{5.60}   & \textbf{4.86}    \\ 
\Xhline{1.1pt}
\multirow{11}{*}{Failure Rate (\%)} & CFSS~\cite{CFSS}  & 20.56           & 66.26           & 23.25           & 17.34           & 21.84           & 32.88           & 23.67            \\
                                   & DVLN~\cite{DVLN}  & 10.84           & 46.93           & 11.15           & 7.31            & 11.65           & 16.30           & 13.71            \\
                                   & LAB~\cite{LAB}                    & 7.56            & 28.83           & 6.37            & 6.73            & 7.77            & 13.72           & 10.74            \\
                                   & SAN~\cite{SAN}                    & 6.32            & 27.91           & 7.01            & 4.87            & 6.31            & 11.28           & 6.60             \\
                                   & WING~\cite{feng2018wing}                   & 6.00            & 22.70           & 4.78            & 4.30            & 7.77            & 12.50           & 7.76             \\ 
                                   & Res-18                 & 10.92           & 43.87           & 13.38           & 7.31            & 11.17           & 16.30           & 11.90            \\
\cline{2-9}
                                   & \textbf{Ours w. Res-18} & 7.44            & 32.52           & 8.60            & 4.30            & 8.25            & 12.77           & 9.06             \\
                                   & \textbf{Ours w. LAB}    & \textbf{5.24}   & \textbf{20.86}  & \textbf{4.78}   & \textbf{3.72}   & \textbf{6.31}   & \textbf{9.51}   & 7.24             \\
                                   & \textbf{Ours w. SAN}    & \textbf{4.08}   & \textbf{18.10}  & \textbf{4.46}   & \textbf{2.72}   & \textbf{4.37}   & \textbf{7.74}   & \textbf{4.40}    \\ 
\Xhline{1.1pt}
\multirow{11}{*}{AUC @0.1}         & CFSS~\cite{CFSS}  & 0.3659          & 0.0632          & 0.3157          & 0.3854          & 0.3691          & 0.2688          & 0.3037           \\
                                   & DVLN~\cite{DVLN}  & 0.4551          & 0.1474          & 0.3889          & 0.4743          & 0.4494          & 0.3794          & 0.3973           \\
                                   & LAB~\cite{LAB}                    & 0.5323          & 0.2345          & 0.4951          & 0.5433          & 0.5394          & 0.4490          & 0.4630           \\
                                   & SAN~\cite{SAN}                    & 0.5355          & 0.2355          & 0.4620          & 0.5552          & 0.5222          & 0.4560          & \textbf{0.4932}  \\
                                   & WING~\cite{feng2018wing}                   & \textbf{0.5504} & \textbf{0.3100} & 0.4959          & 0.5408          & \textbf{0.5582} & \textbf{0.4885} & 0.4918           \\ 
                                   & Res-18                 & 0.4385          & 0.1527          & 0.3718          & 0.4559          & 0.4366          & 0.3655          & 0.3931           \\
\cline{2-9}
                                   & \textbf{Ours w. Res-18} & 0.5034          & 0.2294          & 0.4534          & 0.5252          & 0.4849          & 0.4318          & 0.4532           \\
                                   & \textbf{Ours w. LAB}    & 0.5460          & 0.2764          & \textbf{0.5098} & \textbf{0.5660} & 0.5349          & 0.4700          & 0.4923           \\
                                   & \textbf{Ours w. SAN}    & \textbf{0.5913} & \textbf{0.3109} & \textbf{0.5490} & \textbf{0.6089} & \textbf{0.5812} & \textbf{0.5164} & \textbf{0.5513}  \\
\Xhline{1.2pt}
\end{tabular}%
}
\end{center}
\vspace{-0.5cm}
\caption{\label{tb_wflw} \small{Evaluation of our approach on WFLW dataset. Top-2 results are highlighted in bold font.}}
\end{table*}

\noindent \textbf{WFLW}~\cite{LAB} dataset is a challenging one, which contains 7,500 faces for training and 2,500 faces for testing, based on WIDER Face~\cite{yang2016wider} with 98 manually annotated landmarks~\cite{LAB}. The dataset is partitioned into 6 subsets according to challenging attribute annotation of large pose, expression, illumination, makeup, occlusion, and blur. 

\noindent\textbf{300W}~\cite{sagonas2013300} provides multiple face datasets including LFPW~\cite{LFPW}, AFW~\cite{AFW}, HELEN~\cite{HELEN}, XM2VTS~\cite{messer1999xm2vtsdb}, and IBUG with 68 automatically-annotated landmarks. Following the protocol used in~\cite{LBF}, 3,148 training images and 689 testing images are used. The testing images include two subsets, where 554 test samples from LFPW and HELEN form the common subset and 135 images from IBUG constitute the challenging subset.

\noindent\textbf{AFLW}~\cite{AFLW} dataset is widely used for benchmarking facial landmark localization. It contains 24,386 in-the-wild faces with a wide range of yaw, pitch and roll angles ($[-120^{\circ}, 120^{\circ}]$ for yaw, $[-90^{\circ}, 90^{\circ}]$ for pitch and roll). Following the widely-adopted protocol~\cite{CFSS,zhu2016unconstrained}, the AFLW-full dataset has 20,000 images for training and 4,386 for testing. It is originally annotated with 19 sparse facial landmarks. To provide a better benchmark for evaluating pose variation and allow cross-dataset evaluation, we re-annotate it with 68 facial landmarks, which follow the common standard in 300W~\cite{sagonas2013300,shen2015first}. Based on the new 68-point annotation, we conduct more precise evaluation. Cross-dataset evaluation is also provided among existing datasets~\cite{LFPW,AFW,HELEN}. 

\noindent\textbf{COFW} dataset~\cite{burgos2013robust} contains 1,345 images for training and 507 images for testing, focusing on occlusion. The whole dataset is originally annotated with 29 landmarks and has been re-annotated with 68 landmarks in~\cite{cofw68} to allow cross-dataset evaluation. We utilize 68 annotated landmarks provided by~\cite{cofw68} to conduct comparison with other approaches. 

\subsection{Experimental Setting}
\paragraph{Evaluation Metrics} 
We evaluate performance of facial landmark detection using normalized landmarks mean error and Cumulative Errors Distribution (CED) curve. For the 300W dataset, we normalize the error using inter-pupil distance. In Table~\ref{tb_300w_fullset}, we also report the NME using inter-ocular distance to compare with algorithms of~\cite{SAN,pifa,rdr,PCD-CNN}, which also use it as the normalizing factor. For other datasets, we follow the protocol used in~\cite{sagonas2013300,MDM} and apply inter-ocular distance for normalization. 

\vspace{-0.12in}
\paragraph{Implementation Details} 
Before training, all images are cropped and resized to $256 \times 256$ using provided bounding boxes. For the detailed conditional variational auto-encoder network structures, we use a two-branch encoder-decoder structure as shown in Fig.~\ref{fig:framework}. We use 6 residual encoder blocks for downsampling the input feature maps, where batch normalization is removed for better synthetic results. The facial landmark detector backbone is substitutable and different detectors are usable to achieve improvement, which we will discuss later. 

For training of the disentangling step, we use Adam~\cite{kingma2014adam} with an initial learning rate of $0.01$, which descends linearly to $0.0001$ with no augmentation. For training of detectors, we first augment each landmark map with $k$ random styles sampled from other face images. The number is set to $8$ if not specially mentioned in experiments. For the detector architecture, a simple baseline network based on ResNet-18~\cite{he2016identity} is chosen by changing the output dimension of the last FC layers to landmark $\times$ 2 to demonstrate the increase brought by style translation. To compare with state-of-the-arts and further validate the effectiveness of our approach, we replace our baseline model with similar structures proposed in~\cite{LAB,SAN}, with the same affine augmentation.

\subsection{Comparison with State-of-the-arts}

\paragraph{WFLW} We evaluate our approach on WFLW~\cite{LAB} dataset. WFLW is a recently proposed challenging dataset with images from in-the-wild environment. We compare algorithm in terms of NME(\%), Failure Rate(\%) and AUC(@0.1) following protocols used in~\cite{LAB}.

The Res-18 baseline receives strong enhancement using synthetic images. To further verify the effectiveness and generality of using style information, we replace the network by two strong baselines~\cite{SAN,LAB} and report the result in Table \ref{tb_wflw}. The light-weight Res-18 is improved by $13.8\%$. By utilizing a stronger baseline, our model achieves $4.39\%$ NME under style-augmented training, outperforms state-of-the-art entries by a large margin. In particular, for the strong baselines, our method also brings $15.9\%$ improvement to SAN~\cite{SAN} model, and $9\%$ boost to LAB~\cite{LAB} from $5.27\%$ NME to $4.76\%$. The elevation is also determined by the model capacity.

\begin{table}[htb]
	\begin{center}
		\resizebox{0.9\columnwidth}{!}{
			\begin{tabular}{c|ccc}
				\Xhline{1.2pt}
				\multirow{2}{*}{Method} & Common & Challenging & \multirow{2}{*}{Fullset} \\
				& Subset & Subset & \\
				\Xhline{1.2pt}
				\multicolumn{4}{c}{Inter-pupil Normalization}  \\ 
				\Xhline{1.2pt}
				SDM~\cite{SDM} & 5.57 & 15.40 & 7.52 \\
				CFAN~\cite{cfan} & 5.50 & 16.78 & 7.69 \\
				ESR~\cite{ESR} & 5.28 & 17.00 & 7.58 \\
				LBF~\cite{LBF} & 4.95 & 11.98 & 6.32 \\
				CFSS~\cite{CFSS} & 4.73 & 9.98 & 5.76 \\
				TCDCN~\cite{TCDCN} & 4.80 & 8.60 & 5.54 \\
				RCN~\cite{RCN} & 4.67 & 8.44 & 5.41 \\
				3DDFA~\cite{3DDFA} & 6.15 & 10.59 & 7.01 \\
				SeqMT~\cite{semi} & 4.84 & 9.93 & 5.74 \\
				RAR~\cite{rar} & 4.12 & 8.35 & 4.94 \\
				TSR~\cite{TSR} & 4.36 & 7.56 & 4.99 \\
				DCFE~\cite{DCFE} & 3.83 & 7.54 & 4.55 \\
				LAB ~\cite{LAB} & 4.20 & 7.41 & 4.92 \\
				Res-18 & 4.53 & 8.41 & 5.30 \\
				\hline
				\textbf{Ours w LAB} & \textbf{4.23} & \textbf{7.32} & \textbf{4.83} \\
				\textbf{Ours w Res-18} & \textbf{3.98} & \textbf{7.21} & \textbf{4.54} \\
				\Xhline{1.2pt}
				\multicolumn{4}{c}{Inter-ocular Normalization}  \\
				\Xhline{1.2pt}
				PIFA~\cite{pifa} & 5.43 & 9.88 & 6.30 \\
				RDR~\cite{rdr} & 5.03 & 8.95 & 5.80 \\
				PCD-CNN~\cite{PCD-CNN} & 3.67 & 7.62 & 4.44 \\
				SAN~\cite{SAN} & 3.34 & 6.60 & 3.98 \\
				\hline
				\textbf{Ours w SAN} & \textbf{3.21} & \textbf{6.49} & \textbf{3.86} \\
				\hline
			\end{tabular}
		}
	\end{center}
	\vspace{-0.5cm}
	\caption{\label{tb_300w_fullset} {Normalized mean error (\%) on 300W common, challenging subset and the full set.}}
	\vspace{-0.3cm}
\end{table}

\vspace{-0.12in}
\paragraph{300W} 
In Table~\ref{tb_300w_fullset}, we report different facial landmark detector performance (in terms of normalized mean error) on 300W dataset. The baseline network follows Res-18 structure. With additional ``style-augmented'' synthetic training samples, our model based on a simple backbone outperforms previous state-of-the-art methods. We also report results of models that are trained on original data, which reflect the performance gain brought by our approach.

Similarly, we replace the baseline model with a state-of-the-art method~\cite{SAN}. Following the same setting, this baseline is also much elevated. Note that the 4-stack LAB~\cite{LAB} and SAN~\cite{SAN} are open-source frameworks. We train the models from scratch, which perform less well than those reported in their original papers. However, our model still yields $1.8\%$ and $3.1\%$ improvement on LAB and SAN respectively, which manifest the consistent benefit when using the ``style-augmented'' strategy.

\begin{figure}[t]
\begin{center}
  \includegraphics[width=\linewidth]{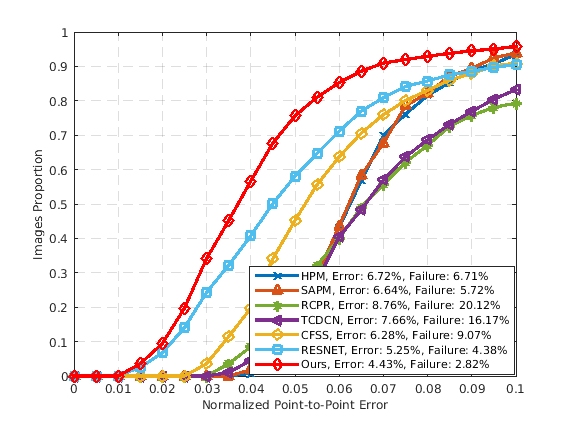}
\end{center}
\vspace{-0.5cm}
   \caption{{Cumulative error distribution curve on COFW 68-point test set.}}
\label{fig:cofw}
\vspace{-0.3cm}
\end{figure}

\begin{figure*}[t]
	\begin{center}
		\includegraphics[width=\linewidth]{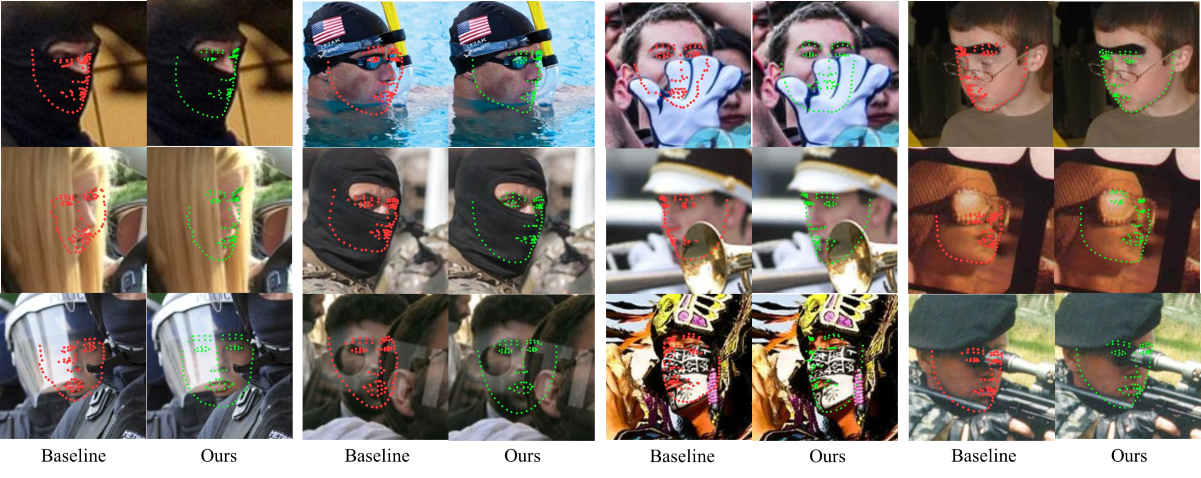}
	\end{center}
	\vspace{-0.5cm}
	\caption{\small{Visual comparison on WFLW test set between the original baseline model and the boosted framework via style translation.}}
	\vspace{-0.2cm}
	\label{fig:comparison}
\end{figure*}

\begin{table}[htb]
	\begin{center}
		\small
		\begin{tabular}{c|cccc}
			\Xhline{1.2pt}
			\multirow{2}{*}{Method} & \multicolumn{2}{c}{NME(\%)} & AUC@ 0.1 & FR(\%) 
			\\ \cline{2-5} 
			& Frontal & Full & Full & Full \\ \hline
			LAB~\cite{LAB} & 2.23 & 7.15 & 0.39 & 11.28\\ 
			SAN~\cite{SAN} & 2.01 & 6.94 & 0.44 & 10.43 \\
			Res-18 & 2.30 & 7.23 & 0.37 & 11.89 \\ 
			\hline
			\hline
			\textbf{Ours w. Res-18} & 2.20 & 7.17 & 0.38 & 11.91 \\
			\textbf{Ours w. LAB}   & 2.10 & 7.06 & 0.42 & 10.01\\ 
			\textbf{Ours w. SAN}   & \textbf{1.86} & \textbf{6.01} & \textbf{0.58} & \textbf{9.70} \\
			\Xhline{1.2pt}
		\end{tabular}
	\end{center}
	\vspace{-0.5cm}
	\caption{\label{tb_aflw} \small{Normalized mean error (\%) on re-annotated 68-pt AFLW frontal subset and the full set.}}
	\vspace{-0.2cm}
\end{table}

\begin{table}[t]
	\begin{center}\small
			\begin{tabular}{ccccc}
				\Xhline{1.2pt}
				\multirow{2}{*}{Dataset} & \multirow{2}{*}{PCT (\%)} & \multicolumn{3}{c}{NME (\%)} \\ \cline{3-5} 
				&  & Res-18 & w Ours & Improved \\ \Xhline{1.2pt}
				\multicolumn{1}{c|}{\multirow{9}{*}{300W}} & 10 & 13.72 & \textbf{7.86} & +42.71\% \\
				\multicolumn{1}{c|}{} & 20 & 9.66 & \textbf{6.07} & +37.16\% \\
				\multicolumn{1}{c|}{} & 30 & 8.9 & \textbf{5.86} & +34.16\% \\
				\multicolumn{1}{c|}{} & 40 & 8.86 & \textbf{5.29} & +40.29\% \\
				\multicolumn{1}{c|}{} & 50 & 7.96 & \textbf{5.23} & +34.30\% \\
				\multicolumn{1}{c|}{} & 60 & 7.89 & \textbf{5.18} & +34.35\% \\
				\multicolumn{1}{c|}{} & 70 & 7.02 & \textbf{5.04} & +28.21\% \\
				\multicolumn{1}{c|}{} & 80 & 6.66 & \textbf{4.82} & +27.63\% \\
				\multicolumn{1}{c|}{} & 90 & 6.58 & \textbf{4.69} & +28.72\% \\ \hline
				\multicolumn{1}{c|}{\multirow{9}{*}{WFLW}} & 10 & 22.09 & \textbf{10.81} & +51.06\% \\
				\multicolumn{1}{c|}{} & 20 & 16.04 & \textbf{8.98} & +44.01\% \\
				\multicolumn{1}{c|}{} & 30 & 13.91 & \textbf{8.24} & +40.76\% \\
				\multicolumn{1}{c|}{} & 40 & 12.19 & \textbf{8.03} & +34.13\% \\
				\multicolumn{1}{c|}{} & 50 & 11.78 & \textbf{7.75} & +34.21\% \\
				\multicolumn{1}{c|}{} & 60 & 10.41 & \textbf{7.31} & +29.78\% \\
				\multicolumn{1}{c|}{} & 70 & 9.87 & \textbf{7.29} & +26.14\% \\
				\multicolumn{1}{c|}{} & 80 & 9.66 & \textbf{7.25} & +24.95\% \\
				\multicolumn{1}{c|}{} & 90 & 9.04 & \textbf{7.19} & +20.46\% \\ \hline
			\end{tabular}
	\end{center}
	\vspace{-0.5cm}
	\caption{\label{tb_ablation_Res} \small{Normalized mean error (\%) on 300W common and WFLW datasets when the training images are split into $10$ folds. Each row represents NME on test set when the model is trained using a percentage (PCT\%) of the training set. The landmark detector backbone is Res-18.}}
\end{table}

\vspace{-0.12in}
\paragraph{Cross-dataset Evaluation on COFW}
To comprehensively evaluate the robustness of our method towards occlusion, COFW-68 is also utilized for cross-dataset evaluation. We perform comparison against several state-of-the-art methods in Fig.~\ref{fig:cofw}. Our model performs the best with $4.43\%$ mean error and $2.82\%$ failure rate, which indicates high robustness to occlusion due to our proper utilization of style translation.

\vspace{-0.12in}
\paragraph{AFLW}
We further evaluate our algorithm on the AFLW~\cite{AFLW} dataset following the AFLW Full protocol. AFLW is also challenging for its large pose variation. It is originally annotated with 19 facial landmarks, which are relatively sparse. To make it more useful, we richen the dataset by re-annotating it with 68-point facial landmarks. This new set of data is also publicly available.

We compare our approach with several models in Table~\ref{tb_aflw}, by re-implementing their algorithms on the new dataset along with our style-augmented samples. Exploiting style information also boosts landmark detectors with a large-scale training set ($25,000$ images in AFLW). Interestingly, our method improves SAN baseline in terms of NME on Full set from $6.94\%$ to $6.01\%$, which indicates that augmenting in style level brings promising improvement on solving large pose variation. The visual comparison in Fig. \ref{fig:comparison} shows hidden face part is better modeled with our strategy.

\subsection{Ablation Study}

\subsubsection{Improvement on Limited Data}
Disentanglement of style and structure is the key that influences quality of style-augmented samples. We evaluate the completeness of disentanglement especially when the training samples are limited. To evaluate the performance and relative gain of our approach when training data is limited. The training set is split into 10 subsets and respectively we evaluate our model on different portions of training data. Note that for different portions, we train the model from scratch with no extra data used. The quantitative result is reported in Tables~\ref{tb_ablation_Res} and \ref{tb_ablation_SAN}.

\begin{table}[t]
\begin{center}
\begin{tabular}{ccccc}
\Xhline{1.2pt}
\multirow{2}{*}{Dataset} & \multirow{2}{*}{PCT (\%)} & \multicolumn{3}{c}{NME (\%)} \\ \cline{3-5} 
 &  & SAN & w Ours & Improved \\ \Xhline{1.2pt}
\multicolumn{1}{c|}{\multirow{9}{*}{300W}} & 10 & 84.33 & \textbf{4.27} & +94.94\% \\
\multicolumn{1}{c|}{} & 20 & 5.08 & \textbf{3.85} & +24.21\% \\
\multicolumn{1}{c|}{} & 30 & 4.05 & \textbf{3.65} & +9.88\% \\
\multicolumn{1}{c|}{} & 40 & 3.8 & \textbf{3.49} & +8.16\% \\
\multicolumn{1}{c|}{} & 50 & 3.6 & \textbf{3.39} & +5.83\% \\
\multicolumn{1}{c|}{} & 60 & 3.54 & \textbf{3.32} & +6.21\% \\
\multicolumn{1}{c|}{} & 70 & 3.48 & \textbf{3.29} & +5.46\% \\
\multicolumn{1}{c|}{} & 80 & 3.39 & \textbf{3.21} & +5.31\% \\
\multicolumn{1}{c|}{} & 90 & 3.38 & \textbf{3.19} & +5.62\% \\ \hline
\multicolumn{1}{c|}{\multirow{9}{*}{WFLW}} & 10 & 9.16 & \textbf{7.2} & +21.40\% \\
\multicolumn{1}{c|}{} & 20 & 7.41 & \textbf{6} & +19.03\% \\
\multicolumn{1}{c|}{} & 30 & 6.73 & \textbf{5.48} & +18.57\% \\
\multicolumn{1}{c|}{} & 40 & 6.26 & \textbf{5.21} & +16.77\% \\
\multicolumn{1}{c|}{} & 50 & 5.95 & \textbf{4.98} & +16.30\% \\
\multicolumn{1}{c|}{} & 60 & 5.72 & \textbf{4.84} & +15.38\% \\
\multicolumn{1}{c|}{} & 70 & 5.5 & \textbf{4.69} & +14.73\% \\
\multicolumn{1}{c|}{} & 80 & 5.43 & \textbf{4.63} & +14.73\% \\
\multicolumn{1}{c|}{} & 90 & 5.23 & \textbf{4.6} & +12.05\% \\ \hline
\end{tabular}
\end{center}
\vspace{-0.5cm}
\caption{\label{tb_ablation_SAN} {Normalized mean error (\%) on 300W common and WFLW datasets when using different percentages of the training set, with the same protocol as in Table~\ref{tb_ablation_Res} on a stronger baseline. The baseline network here follows SAN~\cite{SAN} structure. }}
\end{table}

In Table~\ref{tb_ablation_Res}, a light baseline network Res-18 is used to show the relative improvement on different training samples. Style-augmented synthetic images improve detectors' performance by a large margin, while the improvement is even larger when the number of training images is quite small. In Table~\ref{tb_ablation_SAN}, a stronger baseline SAN~\cite{SAN} is chosen. Surprisingly, the baseline easily reaches state-of-the-art performance using only $50\%$ labeled images, compared to former methods provided in Table~\ref{tb_wflw}.

\begin{figure}[tb]
\begin{center}
\includegraphics[width=\linewidth]{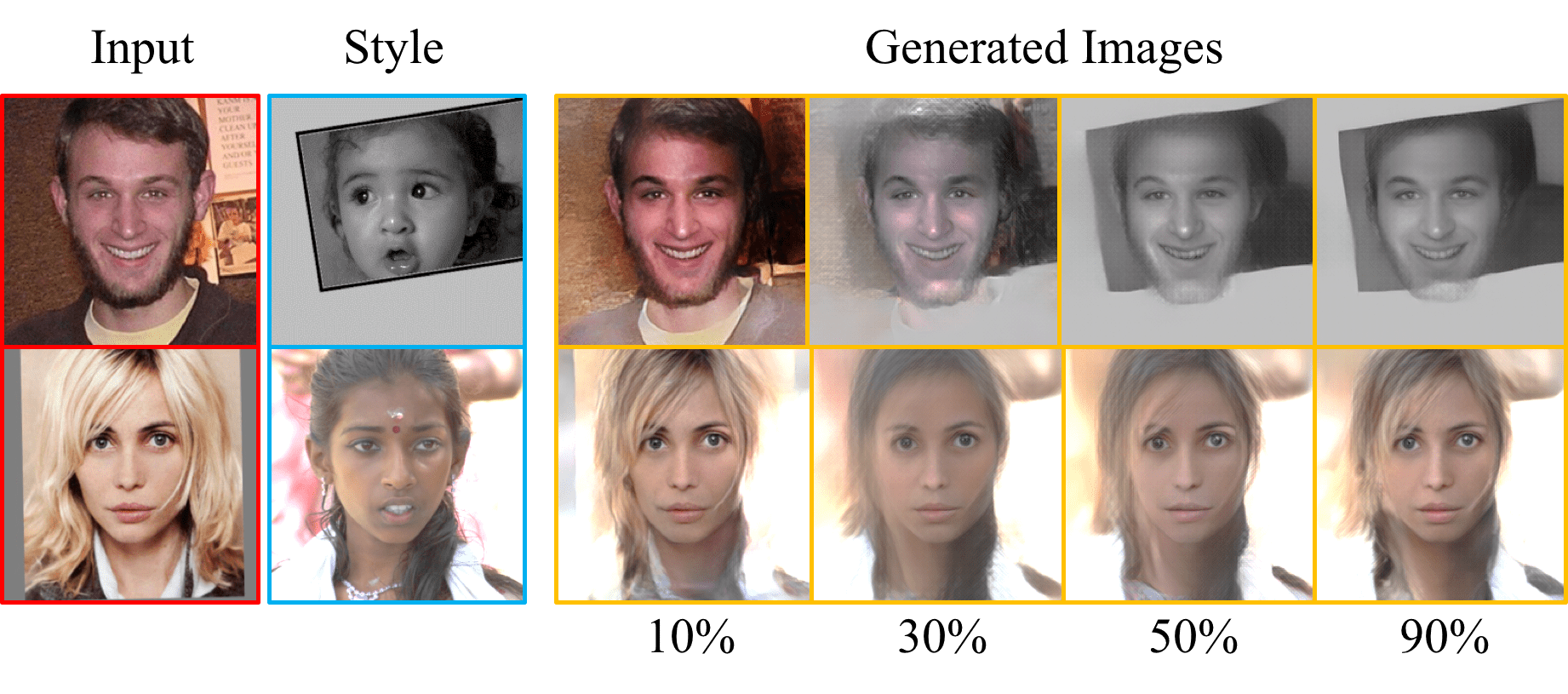}
\end{center}
\vspace{-0.5cm}
\caption{\label{ablation}{Results of style translation using different numbers of data. The left $2$ images are the input, with $2$ different reference styles. The percentage refers to how much data is used to train the disentangle module.}}
\end{figure}

Besides, Fig.~\ref{ablation} provides an intuitive visualization of the resulting generated faces when part of the data is used. Each column contains output that is rendered from the input structure and given style, when using a portion of face image data. It shows when the data is limited, our separation component tends to capture weak style information, such as color and lighting. Given more data as examples, the style becomes complex and captures detailed texture and degradation, like occlusion. 

The results verify that even using limited labeled images, our design is capable of disentangling style information and keeps improve those baseline methods that are already very strong.

\begin{table}[tb]
\begin{center}
\begin{tabular}{c|c|c|c|c|c|c}
\Xhline{1.2pt}
Number  & 0    & 2    & 4    & 8    & 16   & 32   \\ \hline
NME (\%) & 6.22 & 5.89 & 5.54 & 5.31 & \textbf{5.29} & 5.34 \\ 
\Xhline{1.2pt}
\end{tabular}
\vspace{-0.3cm}
\end{center}
\caption{\label{stlye_num} {Normalized mean error (\%) on WFLW test set using different numbers of style translation.}}
\vspace{-0.3cm}
\end{table}

\subsubsection{Estimating the Upper-bound}
As discussed before, our method conceptually and empirically augments training with $n^2$ synthetic samples. By augmenting each face image with $k$ random styles, the training set could be very large and slows down convergence. In this section, we experiment with choosing the style augmenting factor $k$ and test the upper bound of style translation. We evaluate our method by adding the number of random sampled styles $k$ of each annotated landmarks on a ResNet-50 baseline. 

The result is reported in Table \ref{stlye_num}. By adding a number of augmented styles, the model continue gaining improvement. However, when $k \ge 8$, the performance grow slows down. It begins to decrease if $k$ reaches $32$. The reason is that due to the quantity imbalance between real and synthetic faces, a very large $k$ makes the model overfit to synthetic image texture when the generated image quantity is large.

\section{Conclusion and Future Work}

In this paper, we have analyzed the well-studied facial landmark detection problem from a new perspective of implicit style and environmental factor separation and utilization. Our approach exploits the disentanglement representation of facial geometry and unconstrained style to provide synthetic faces via style translation, further boosting quality of facial landmarks. Extensive experimental results manifest its effectiveness and superiority. 

We also note that utilizing synthetic data for more high-level vision tasks still remains an open problem, mainly due to the large domain gap between generated and real images. In our future work, we plan to model style in a more realistic way by taking into account the detailed degradation types and visual quality. We also plan to generalize our structure to other vision tasks.

{\small
\bibliographystyle{ieee_fullname}
\bibliography{egbib}
}

\clearpage
\section*{Appendix}
The content of our supplementary material is organized as follows.

\begin{enumerate}
    \item More ablation studies and detailed analysis of components in our framework. 
    \item Additional discussion about related directions.
    \item Details of our annotated AFLW-68 dataset and some representative visualized samples.
\end{enumerate}

\subsection*{S1. More Ablation Studies}
In this section, we provide additional analysis about each design in our framework to facilitate understanding of our structure. Two key loss terms in our framework are studied to give insights into their respective roles. Qualitative visualization and quantitative results are reported for a comprehensive comparison.

\begin{figure}[htb]
	\begin{center}
		\includegraphics[width=\linewidth]{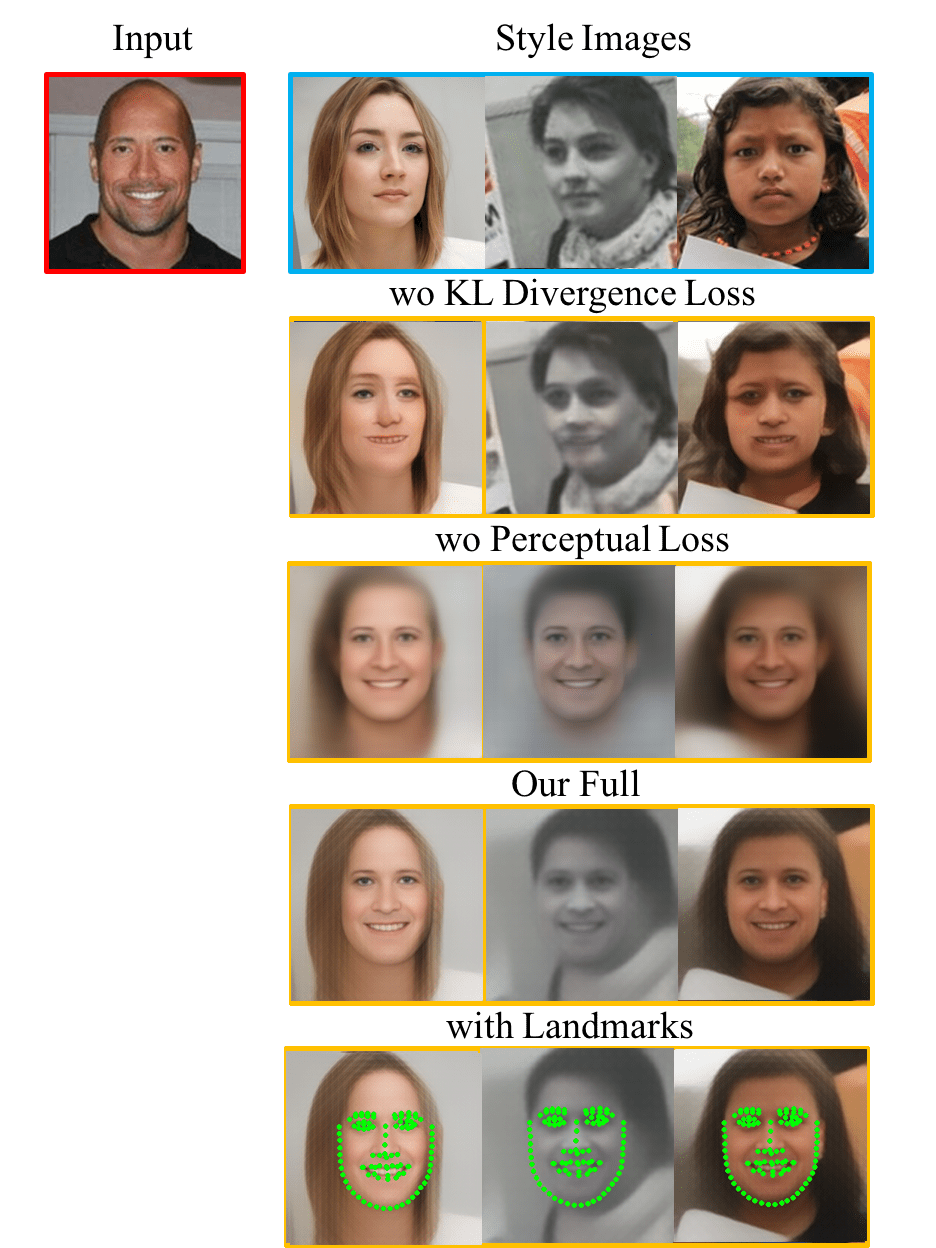}
	\end{center}
	\vspace{-0.5cm}
	\caption{\small{Qualitative analysis of each component in our framework. Given input images in red, $3$ different styles are provided to perform translation towards input structure. $3$ incomplete variants of our framework are used to show the functionality of each component.}}
	\vspace{-0.2cm}
	\label{fig:ablation_vis}
\end{figure}

KL divergence loss and perceptual loss, are incorporated into our framework during the disentangled learning procedure. Fig.~\ref{fig:ablation_vis} shows their respective effect on style translation via visual comparisons of several incomplete variants. Through visual observations, their roles could be inferred intuitively. The perceptual loss, as discussed, is designed to capture better style information and visual quality. Thus, removing this term leads to ``over-smoothness'' and poor diversity on synthetic images. Removing KL divergence term shows severe structure distortion on translated results, which indicates that KL divergence loss plays a key role on disentangling structure and style information. 

Quantitative results of each variants are also reported in Table.~\ref{ablation_quantitive}. The normalized mean error(NME) is evaluated on WFLW~\cite{LAB} test set when the model is trained on style augmented dataset using each variant. We observe that NME will increase if any loss function is removed. In particular, the detector performance drops significantly lower than the baseline if $L_{KL}$ is removed. Both the qualitative and quantitative result interprets the role of each component, indicating their essentialness in our framework. 

\begin{table}[htb]
\begin{center}
\resizebox{0.9\columnwidth}{!}{%
\begin{tabular}{c|c|c|c|c}
\Xhline{1.2pt}
Model  & Baseline    & wo KL divergence     & wo Perceptual    & Full   \\ \hline
NME(\%) & 8.49 & 9.08 & 8.34 & \textbf{7.98} \\ 
\Xhline{1.2pt}
\end{tabular}
}
\vspace{-0.3cm}
\end{center}
\caption{\label{ablation_quantitive} \small{Quantitative ablative results. Normalized mean error (\%) on WFLW test set using different variants of our framework.}}
\vspace{-0.3cm}
\end{table}

\subsection*{S2. Additional Discussion }

In this section, we provide more discussion on our approach， along with our analysis towards some existing alternatives.

\subsubsection*{S2.1 Comparison with GAN-based approaches}
Generative adversarial network~(GAN) and its applications are widely studied these days, using GAN-synthetic data to aid training, has also been explored along this line. Some works~\cite{antoniou2018data} have utilized GANs to perform data augmentation. However, its effect still remains questionable especially on high-level vision challenges. For instance, in our task, face images need to be labeled with accurate landmarks. Existing generative models are incapable of handling these  tasks with fine-grained annotations, e.g. semantic segmentation, constrained by its limited generalizability. We choose to escape the difficulties of GAN training, starting from a new perspective of internal representation. With decent representation of separating style and structure, different interactions within a face image can be simulated by re-rendering from existing style and structure code. In other words, our choice depends upon fully exploiting available information by mixing them, instead of creating new information and visually perfect results via adversarial learning procedure. However, if two codes of structure and style are factored well, advances on high fidelity images synthesis could theoretically bring more gains based on our framework. 

\subsubsection*{S2.2 Comparison with Style Transfer}

\begin{figure}[htb]
	\begin{center}
		\includegraphics[width=\linewidth]{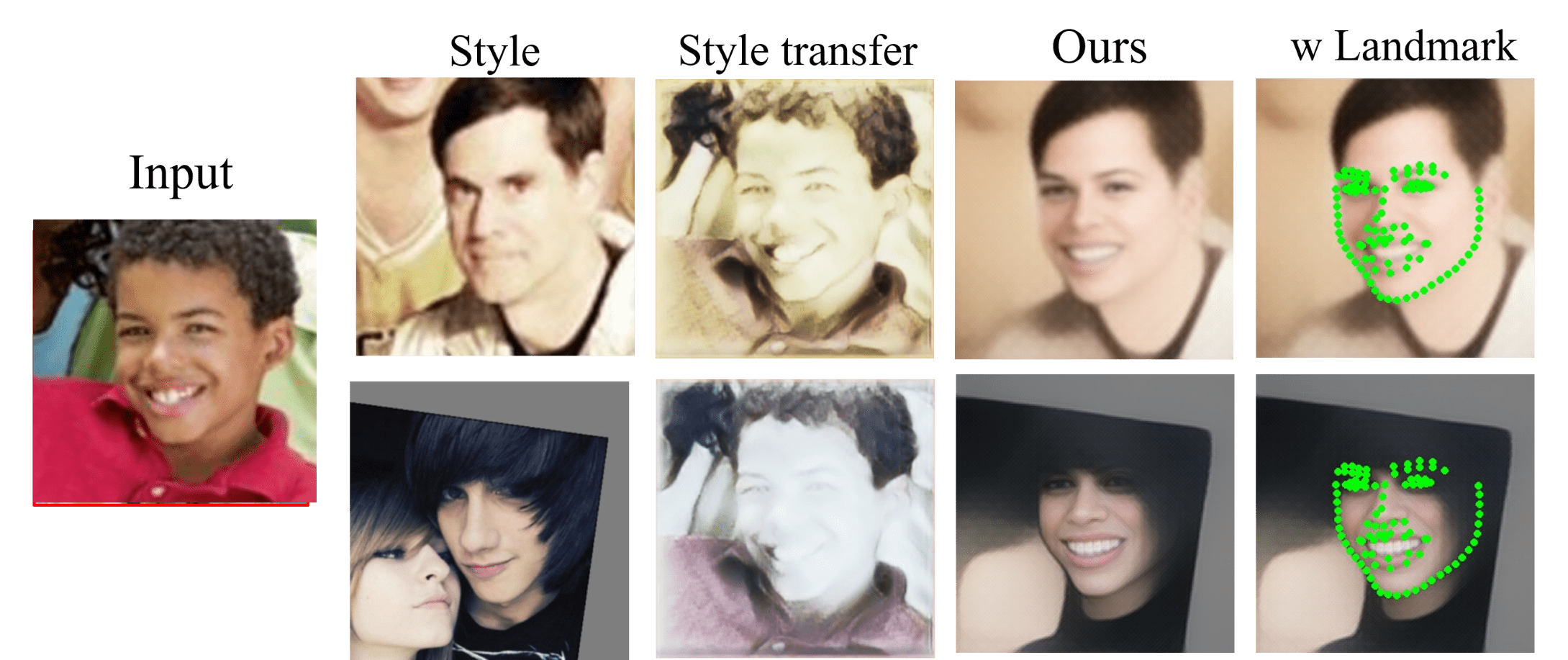}
	\end{center}
	\vspace{-0.5cm}
	\caption{\small{Visual comparison with style transfer approach. For the style transfer algorithm, we use~\cite{gatys2016image}. Our results are more realistic than stylized images, with better structure coherence.}}
	\vspace{-0.2cm}
	\label{fig:style_trans}
\end{figure}

Our method is motivated by advances in style transfer. A common doubt could be why not directly conducting style transfer as a augmentation or how basic style transfer could help training. As discussed, our definition of style includes environments and degradation that prevent the model from recognizing while content refers to facial geometry. Applying ``vanilla" style transfer would leads to structural distortion on stylized images, as illustrated in Fig.~\ref{fig:style_trans}.Our definition of ``style'' helps preserve structure on synthetic images. Besides, synthetic images using style transfer have a large domain gap with real-world face images. Simply augmenting training with these samples would instead hurt model's localization ability on real images.

\begin{table}[htb]
\begin{center}
\begin{tabular}{l|c|l}
\Xhline{1.1pt}
Database  & \multicolumn{1}{l|}{Environment} & Number \\ \Xhline{1.1pt}
Multi-PIE~\cite{gross2010multi} & \multirow{3}{*}{Controlled}      & 750000 \\ \cline{1-1} \cline{3-3} 
XM2VTS~\cite{messer1999xm2vtsdb}    &                                  & 2360   \\ \cline{1-1} \cline{3-3} 
LFPW~\cite{LFPW}     &                                  & 1035   \\ \Xhline{1.1pt}
HELEN~\cite{HELEN}     & \multirow{5}{*}{In-the-wild}     & 2330   \\ \cline{1-1} \cline{3-3} 
AFW~\cite{AFW}       &                                  & 468    \\ \cline{1-1} \cline{3-3} 
IBUG      &                                  & 135    \\ \cline{1-1} \cline{3-3} 
COFW-68~\cite{cofw68}   &                                  & 507    \\ \cline{1-1} \cline{3-3} 
\textbf{AFLW-68(Ours)}   &                                  & \textbf{25993}  \\ \Xhline{1.1pt}
\end{tabular}
\caption{\label{68pts} \small{Widely-used 68-pt facial landmark datasets. Dataset names their the environment and number are reported.}}
\end{center}
\end{table}

\subsection*{S3. Details of AFLW 68-point dataset}

\begin{figure}[htb]
	\begin{center}
		\includegraphics[width=\linewidth]{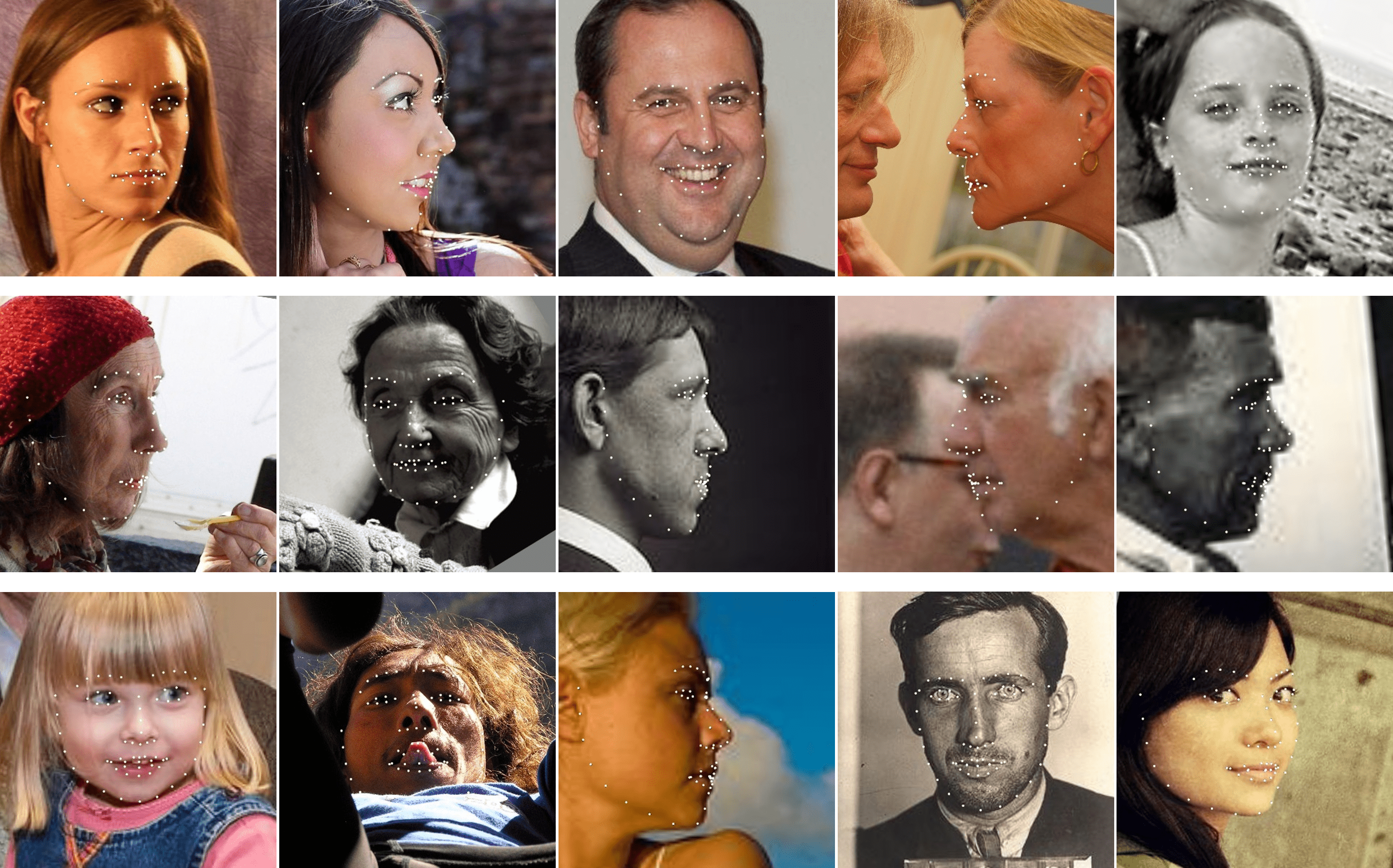}
	\end{center}
	\vspace{-0.5cm}
	\caption{\small{Sampled annotated images in the proposed AFLW 68-point dataset, including in-the-wild faces under large pose variations}}
	\vspace{-0.2cm}
	\label{fig:aflw}
\end{figure}

We propose a new facial landmark dataset based on AFLW~\cite{AFLW}, to facilitate benchmarking on large pose performance. To allow a more precise evaluation and cross-dataset comparison, we follow the widely-used Multi-PIE~\cite{gross2010multi} and 300W~\cite{sagonas2013300} 68-point protocol. Annotated samples are provided at Fig.~\ref{fig:aflw}, which contains extreme pose variations.

\end{document}